\documentclass[letterpaper, 10 pt, conference]{ieeeconf}

\IEEEoverridecommandlockouts %
\newcommand{\mixeropt}{{MIXER}\textsuperscript{*}}
\newcommand{\mixer}{{MIXER}}

\usepackage{amsmath} %
\usepackage{amssymb} %
\usepackage{amsfonts}
\usepackage[usenames,dvipsnames]{xcolor}
\usepackage{tabularx}
\usepackage{multirow}
\usepackage{mathtools}
\usepackage{esvect} %
\usepackage[pagebackref=true,breaklinks=true,colorlinks,bookmarks=false]{hyperref}
\usepackage[binary-units=true]{siunitx}
\usepackage{etoolbox}
\usepackage[normalem]{ulem}
\robustify\bfseries
\robustify\uline

\usepackage{url}

\usepackage{textcomp}

\usepackage[T1]{fontenc} %
\usepackage{subcaption}

\usepackage{balance} %
\usepackage{booktabs} %
\usepackage[export]{adjustbox} %

\newcolumntype{x}[1]{%
>{\raggedleft\hspace{0pt}}p{#1}}%

\usepackage{algorithm}
\usepackage[noend]{algpseudocode} %
\definecolor{commentclr}{RGB}{34, 139, 34}

\newcommand{\ra}[1]{\renewcommand{\arraystretch}{#1}}

\usepackage{amsthm}

\theoremstyle{definition}

\usepackage{letltxmacro}
\LetLtxMacro\orgvdots\vdots
\LetLtxMacro\orgddots\ddots

\makeatletter
\DeclareRobustCommand\vdots{%
	\mathpalette\@vdots{}%
}
\newcommand*{\@vdots}[2]{%
	\sbox0{$#1\cdotp\cdotp\cdotp\m@th$}%
	\sbox2{$#1.\m@th$}%
	\vbox{%
		\dimen@=\wd0 %
		\advance\dimen@ -3\ht2 %
		\kern.5\dimen@
		\dimen@=\wd2 %
		\advance\dimen@ -\ht2 %
		\dimen2=\wd0 %
		\advance\dimen2 -\dimen@
		\vbox to \dimen2{%
			\offinterlineskip
			\copy2 \vfill\copy2 \vfill\copy2 %
		}%
	}%
}
\DeclareRobustCommand\ddots{%
	\mathinner{%
		\mathpalette\@ddots{}%
		\mkern\thinmuskip
	}%
}
\newcommand*{\@ddots}[2]{%
	\sbox0{$#1\cdotp\cdotp\cdotp\m@th$}%
	\sbox2{$#1.\m@th$}%
	\vbox{%
		\dimen@=\wd0 %
		\advance\dimen@ -3\ht2 %
		\kern.5\dimen@
		\dimen@=\wd2 %
		\advance\dimen@ -\ht2 %
		\dimen2=\wd0 %
		\advance\dimen2 -\dimen@
		\vbox to \dimen2{%
			\offinterlineskip
			\hbox{$#1\mathpunct{.}\m@th$}%
			\vfill
			\hbox{$#1\mathpunct{\kern\wd2}\mathpunct{.}\m@th$}%
			\vfill
			\hbox{$#1\mathpunct{\kern\wd2}\mathpunct{\kern\wd2}\mathpunct{.}\m@th$}%
		}%
	}%
}
\makeatother

\let\oldnl\nl%
\newcommand{\nonl}{\renewcommand{\nl}{\let\nl\oldnl}}%
\makeatother

\def\br{\mathbb R}

\def\sa{\mathcal A}

\def\su{\mathcal U}

\def\sl{\mathcal{L}}

\def\ie{i.e., }
\def\eg{e.g., }

\newcommand\eqdef{\mathrel{\overset{\makebox[0pt]{\mbox{\normalfont\tiny def}}}{=}}}

\title{\Large \bf
MIXER: A Principled Framework for Multimodal, Multiway Data Association
	}

\author{Parker C. Lusk*, Ronak Roy*, Kaveh Fathian*, Jonathan P. How%
	\thanks{P.\ C.\ Lusk, R. Roy, K.\ Fathian and J.\ P.\ How are with the Department of Aeronautics and Astronautics, Massachusetts Institute of Technology.
	    {\{plusk, ronakroy, kavehf, jhow\}@mit.edu.} *Authors contributed equally.}
    \thanks{This work is supported by ARL DCIST under Cooperative
    Agreement Number W911NF-17-2-0181.}
}%

\let\oldtwocolumn\twocolumn
\renewcommand\twocolumn[1][]{%
    \oldtwocolumn[{#1}{
    \begin{center}
        \vspace{-0.75cm}
		\includegraphics[width=1\linewidth]{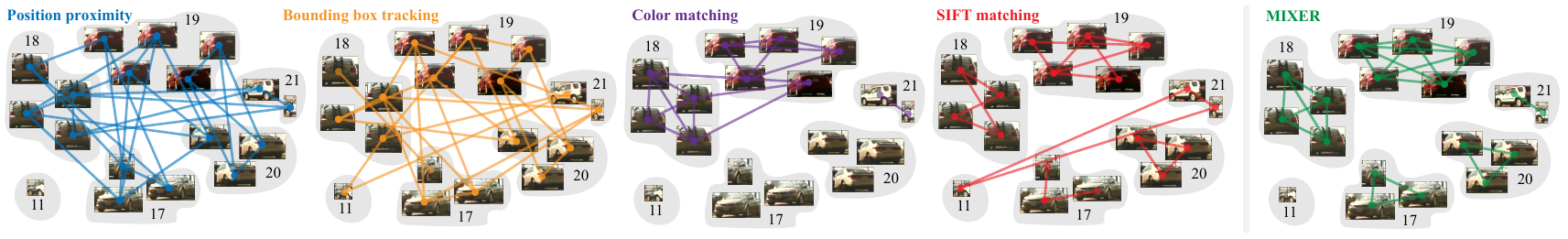}
        \captionof{figure}{
        Small-scale demonstration of MIXER associating identical cars.
        Each labeled cluster (grey) represents multiple observations of the same car.
        Based on four distinct matching modalities of position proximity, bounding box tracking, color, and SIFT matching, observations are associated if they are sufficiently similar.
        While each modality has erroneous associations (\ie observations of different cars are thought to be of the same car), MIXER combines these modalities to yield a fully-accurate set of associations, correctly grouping all observations of the same car.
        }
    	\vspace{-0.1cm}
    	\label{fig:teaser}
        \end{center}
    }]
}

\begin{document}

\maketitle

\thispagestyle{plain}
\pagestyle{plain}

\begin{abstract}
A fundamental problem in robotic perception is matching identical objects or data, with applications such as loop closure detection, place recognition, object tracking, and map fusion. 
While the problem becomes considerably more challenging when matching should be done jointly across multiple, multimodal sets of data, the robustness and accuracy of matching in the presence of noise and outliers can be greatly improved in this setting. 
At present, multimodal techniques do not leverage multiway information, and multiway techniques do not incorporate different modalities, leading to inferior results. 
In contrast, we present a principled mixed-integer quadratic framework to address this issue.   
We use a novel continuous relaxation in a projected gradient descent algorithm that guarantees feasible solutions of the integer program are obtained efficiently.
We demonstrate experimentally that correspondences obtained from our approach are more stable to noise and errors than state-of-the-art techniques.
Tested on a robotics dataset, our algorithm resulted in a $35\%$ increase in $\mathrm{F}_1$ score when compared to the best alternative.

\end{abstract}

\section{Introduction}\label{sec:intro}

Identifying correspondences across sets of data, or across sources with different modalities, is a fundamental problem in robotics and computer vision.
In practice, data points are noisy and contain outliers that should not be matched. 
These challenges make classical assignment techniques such as the Hungarian \cite{kuhn1955hungarian} or auction \cite{bertsekas1988auction} algorithms ineffective, as they cannot reject outliers or produce consistent results across sets. 
State-of-the-art multiway data association algorithms \cite{Pachauri2013, Maset2017, Leonardos2017, Zhou2015, Bernard2018, tron2017quickmatch, fathian2020clear} can remove outliers and produce consistent associations by matching data jointly across all sets, however, these techniques cannot fuse associations that come from different modalities. Furthermore, these schemes operate based on binary associations, \ie data points should either be matched or not. Considering lack of information (\ie when correspondences cannot be established and the decision should be delayed) is an important feature that is not currently present.
In this work, we present the \textbf{M}ultimodality association matr\textbf{IX} fus\textbf{ER} (MIXER) algorithm. 
MIXER is a principled framework for associating data that contains outliers, comes from different modalities, and is observed across multiple instances.
The MIXER formulation fuses modalities and allows incorporating uncertain or missing information in the decision making process.
This is crucial if a modality is not applicable for establishing correspondences at a certain time.
The associations returned by MIXER are guaranteed to be consistent across all sets and respect additional constraints imposed on the problem.
A small-scale demonstration is shown in Fig.~\ref{fig:teaser}, where MIXER recovers $100\%$ precision with $100\%$ recall from the noisy input. %

We formulate the problem as a mixed-integer quadratic program (MIQP).
Since this MIQP is not scalable to large-sized problems, we present a continuous relaxation to efficiently obtain approximate solutions.
The main contribution of our approach over similar relaxation techniques used in the literature is that solutions of the relaxed problem are guaranteed to converge to feasible, binary solutions of the original problem.
Thus, rounding results to binary values, which is required when using other techniques and may lead to infeasible solutions, is avoided. 
To solve the relaxed problem efficiently, we present a projected gradient descent algorithm with backtracking line search based on the Armijo procedure. 
This polynomial-time algorithm has worst case cubic complexity in problem size (from matrix-vector multiplications) at each iteration, and is guaranteed to converge to stationary points. 
Proofs are omitted due to space limitations, but will be provided in future work.

We evaluate MIXER on both synthetic and real-world datasets and compare the results with state-of-the-art multiway data association algorithms.
Our synthetic analysis demonstrates a small optimality gap between MIXER's solution and the global minimum of the MIQP (max $3.3\%$), while achieving an average runtime of \SI{252}{\milli\second}---an average speedup of $685\times$ over the MIQP solver.
Our real-world evaluation considers associating identical cars observed by a robot moving in a parking lot.
We use the four distinct modalities of proximity, color, image features, and bounding box tracks for associating cars. 
Benchmarking MIXER against the state-of-the-art shows superior accuracy on individual modalities, and further shows that MIXER is effectively able to combine all similarity scores, improving the $\mathrm{F}_1$ score by $35\%$ compared to the best competing algorithm.
In summary, the main contributions of this work include:
\begin{itemize}
\itemsep0em
    \item A novel and principled MIQP framework for multimodal, multiway data association  
    \item A continuous relaxation of the MIQP leading to feasible, binary solutions that can be computed efficiently.
    \item A polynomial-time algorithm for solving the relaxed problem based on projected gradient descent and Armijo procedure with convergence guarantees.
    \item Improvements over state-of-the-art multiway data association algorithms showcased on a realworld dataset with four distinct modalities. 
\end{itemize}
We expect MIXER to significantly improve the accuracy and robustness of existing data association techniques used in robotics and computer vision  applications such as feature matching \cite{Maset2017}, 
multiple object tracking~\cite{Luo2021motlitrev}, person re-identification (ReID)~\cite{karanam2019reid,Zheng2020heteroreid}, place recognition~\cite{Lowry2016visualplacerecognition}, and loop closure detection~\cite{qin2021semantic}.

\subsection{Related Work}

Associating elements from two sets is traditionally formulated as a linear assignment problem~\cite{burkard2009assignment,munkres1957algorithms}, which can be solved in polynomial time~\cite{kuhn1955hungarian,bertsekas1988auction}.
If elements have underlying structure (\eg geometry) that should be considered, the problem can be formulated as a quadratic assignment program \cite{koopmans1957assignment, lawler1963quadratic, loiola2007survey}.
Unlike linear assignment, quadratic assignment (or its MIQP graph matching formulation) is, in general, NP-hard \cite{sahni1976p}.
Exact methods for solving quadratic assignment use expensive branch and bound techniques \cite{bazaraa1979exact, bazaraa1983branch}.
More efficient, but approximate solutions can be obtained, for example, from spectral relaxations \cite{leordeanu2005spectral},
linear relaxations \cite{swoboda2017study},
and convex relaxations \cite{zhao1998semidefinite, NIPS2013_647bba34, kezurer2015tight, bernard2018ds}. 

\textbf{Multiway association.}
Multiway data association frameworks jointly associate elements across multiple sets to ensure (cycle) consistency of associations. 
This can be formulated as permutation synchronization \cite{Pachauri2013}, which is computationally challenging due to binary constraints. 
With the exception of (expensive) combinatorial methods \cite{Zach2010,Nguyen2011}, existing works focus on relaxations to obtain approximate answers; for example,
spectral relaxation \cite{Pachauri2013, Maset2017},
convex relaxation \cite{Chen2014, Hu2018, Yu2016, Leonardos2017, Leonardos2018a},
matrix factorization \cite{Zhou2015, Bernard2018, leonardos2020low},
and graph clustering \cite{Yan2016a, tron2017quickmatch, serlin2020distributed, fathian2020clear}.
While some of these methods accept weighted inputs, they are formulated for synchronizing unimodal, binary associations. 
Lastly, when data has underlying structure, the formulation becomes a multi-graph matching problem \cite{yan2013joint, shi2016tensor, Yan2016a, Swoboda2019}, which is considerably more computationally demanding.

\textbf{Multimodal association.}
Fusing multiple modalities for data association~\cite{durrantwhyte1990sensor,durrantwhyte2016multisensor,castanedo2013review,bar1993estimation} can be seen as a classifier combination problem~\cite{kittler1998combining,kuncheva2014combining}.
Methods of combining classifiers include
evidential reasoning~\cite{shafer1976mathematical,sentz2002combination,murphy1998dempster},
Bayesian methods~\cite{cheeseman1985defense},
and deep learning approaches~\cite{liu2018learn}.
A popular and foundational approach to combine classifiers is AdaBoost~\cite{freund1996adaboost}, which has been used to combine different types of LiDAR features for loop closure detection~\cite{granstrom2011learning}.

\section{MIXER Formulation}\label{sec:formulation}

Consider $n$ sets of data $\sa_i,\, i = 1, \dots, n$, with cardinality $|\sa_i|=m_i$ and $m = \sum_{i =1}^{n}{m_i}$.
We define the \textit{universe} as $\su \eqdef \cup_i  \,\sa_i$, with $|\su|=m_u \le m$ the number of distinct elements across all sets.
For each of the $l$ modalities which associate identical elements across sets, a scalar \textit{similarity score} $s \in [0,1]$ is produced.
Scores of $1$, $0.5$, and $0$ correspond to maximum similarity, lack of information/preference, and maximum dissimilarity, respectively.
We arrange these scores in diagonal \textit{score matrices} defined as $S=\mathrm{diag}(s_1,\dots,s_l)\in[0,1]^{l\times l}$.

Given two sets $\sa_i$ and $\sa_j$,
we define the \textit{association matrix} between the elements of these sets as 
\begin{gather} \label{eq:Aij}  \scriptstyle
A_{ij}\;\eqdef\;\begin{bsmallmatrix}
S_{1 1} & \cdots & S_{1 m_j} \\
\vdots & \ddots & \vdots \\
S_{m_i 1} & \cdots & S_{m_i m_j}
\end{bsmallmatrix} \in [0,1]^{m_i l \times m_j l},
\end{gather}
where $S_{ab}$ denotes the score matrix between elements $a \in \sa_i$ and $b \in \sa_j$. 
These pairwise $A_{ij}$ are used to create the symmetric \textit{aggregate association matrix} between all sets as
\begin{gather} \label{eq:A} \scriptstyle
A\;\eqdef\; \begin{bsmallmatrix}
    A_{11} & \cdots & A_{1n} \\
    \vdots  &  \ddots & \vdots  \\
    A_{n1} & \cdots & A_{nn} 
\end{bsmallmatrix} \in  [0,1]^{m l \times m l}.
\end{gather}

\textbf{Ground-truth association.}
In the ideal setting, score matrices $S$ are either identity or zero and can be compactly represented using Kronecker products as $1 \otimes I_l$ or $0 \otimes I_l$, where $I_l$ is an $l\times l$ identity matrix.
Furthermore, matrix $A$ can be factorized as ${A = U\, U^\top \otimes I_l}$, where 
\begin{gather} \label{eq:U} \scriptstyle
U^\top\;\eqdef\; \begin{bsmallmatrix}
    U_{1}^\top  &
    \dots &
    U_{n}^\top  
\end{bsmallmatrix} \in  \{0 , 1 \}^{m_u \times m}.
\end{gather}
Matrices $U_i \in  \{0 , 1 \}^{m_i \times m_u}$ represent associations between elements of $A_i$ and the universe $\su$.

\textbf{Constraints.}
Often, data association algorithms must meet certain constraints imposed by the high-level task.
The \textit{one-to-one} constraint states that an object cannot be associated with other objects and is satisfied if each row of $U$ has a single $1$ entry. 
The \textit{distinctness} constraint states that objects within a set are distinct and therefore should not be associated.
This is satisfied if there is at most a single $1$ entry in each column of $U_i$.
When the association problem is solved across more than two sets, it is important to ensure associations are \textit{cycle consistent}, which states that if $a \sim b$ and $b \sim c$, then $a \sim c$ and is satisfied if the association matrix can be factorized as $A = U \, U^\top \otimes I_l$. This fact is proven in \cite{tron2017quickmatch} for associations of single modality ($l = 1$), which generalizes similarly to the multimodal case ($l \geq 2$).
These three constraints are crucial for detecting and correcting erroneous similarity scores and associations, but increase the difficulty and complexity of the data association procedure.

\textbf{Optimization problem.}
In practice, similarity scores are noisy and can lead to incorrect associations.
Therefore, our objective is to map these scores to their closest binary value while respecting/leveraging the one-to-one, distinctness, and cycle consistency constraints to correct potential scoring mistakes.     
This goal can be formally stated as finding $U$ that solves the MIQP with Frobenius objective
\begin{gather} \label{eq:frobenius}
\begin{aligned}
& \underset{U \in \{0, 1\}^{m\times m}}{\text{minimize}} & &  \big\| U\, U^\top  \otimes I_l -  A \big\|_{F}^2 ~  {\color{gray} \text{(cycle consistency)}}  \\
& ~~\text{subject to}  & &  U \, \mathbf{1}_m  = \mathbf{1}_m  \quad  ~ \,  {\color{gray} \text{(one-to-one constraint)}} \\
&&&  U_i^\top \mathbf{1}_{m_i} \leq \mathbf{1}_{m_i} ~  {\color{gray} \text{(distinctness constraint)}}
\end{aligned}
\end{gather}
where $\mathbf{1}$ denotes a vector of ones.
We note that the size of $U$ in \eqref{eq:frobenius} is $m \times m$ as opposed to $m \times m_u$ defined in \eqref{eq:U}.
This is because the number of unique elements across all sets (size of universe, $m_u$) is unknown a priori.
A solution $U$ of \eqref{eq:frobenius} has only $m_u$ nonzero columns.

\section{Continuous Relaxation and Algorithm}\label{sec:theory}

Solving \eqref{eq:frobenius} to global optimality becomes impractical as the problem size grows; hence, we propose a relaxation. %
Existing relaxation techniques require rounding, which may produce infeasible solutions.
A key contribution of our approach is that solutions are guaranteed to converge to feasible, binary solutions of the original problem without rounding.

Manipulation of the objective of \eqref{eq:frobenius} and the use of penalty functions to incorporate the constraints into the objective yields the continuous relaxation over the nonnegative reals
\begin{gather} \label{eq:relaxed}
\begin{aligned}
& \underset{U \in \br_{+}^{m\times m}}{\text{minimize}} & &  \big < U U^\top,  \bar{A} \big > +  d \, \big( \big < U^\top U, P_o \big > + \big < U U^\top, P_d \big >   \\[-0.3cm]
&&& \qquad + \big< U^\top U - U^\top - U ,~ \mathbf{1}_{m\times m} \big> \big) \\
& \text{subject to}  & &  U \mathbf{1}_{m} - \mathbf{1}_{m} \leq \mathbf{0}_{m}
\end{aligned}
\end{gather}
where penalty matrices $P_o$ and $P_d$ correspond to orthogonality and distinctness constraints,
and $d \geq 0$ is a scalar parameter.
As $d$ increases, the positive penalty value pushes the solution $U$ toward having orthogonal columns, satisfying the distinctness constraint, and having row-sum equal to $1$.  
Once $d$ is large enough, solutions of \eqref{eq:relaxed} become binary and satisfy the constraints of the original problem \eqref{eq:frobenius}. 
Note that $\bar{A}$, defined in the expansion of objective of \eqref{eq:frobenius}, can be indefinite, making \eqref{eq:relaxed} non-convex in general.

Problem \eqref{eq:relaxed} is efficiently solved using a first-order projected gradient descent scheme with greedy steps and backtracking Armijo line search \cite{bertsekas1997nonlinear}.
The worst case complexity of this local search is bounded by $\mathcal{O}(m^3)$ per iteration, corresponding to matrix multiplications.

\section{Experiments}\label{sec:experiments}
We evaluate MIXER on two datasets.
First, we use synthetic data and study the optimality gap of MIXER, finding that over the problem sizes considered, MIXER can achieve near-optimal performance with significantly improved runtime.
Then, we demonstrate the ability of MIXER to combine sensing modalities on a challenging robotics dataset collected as part of this work.
We compare the performance of MIXER on this dataset with other multiway matching algorithms.

\textbf{Synthetic Dataset.}
We use synthetically generated data to empirically analyze MIXER's optimality gap and runtime.
Data is generated using partial views of $m_u=10$ objects with randomly added outliers and noise, resulting in a problem sizes of $10\le m \le 50$.
We use Gurobi 9.1.1~\cite{gurobi} to solve problem~\eqref{eq:frobenius} to optimality and refer to this implementation as \mixeropt.
MIXER is executed in MATLAB on an i7-6700. %

Table~\ref{tbl:optimality_gap} shows the percent change of the objective value, precision, and recall relative to \mixeropt, absolute runtime of \mixer\ in milliseconds, and the relative speedup of \mixer.
\mixer\ is able to achieve a strikingly small optimality gap, while gaining a considerable speedup, with absolute runtimes less than \SI{300}{\milli\second}.
Interestingly, we observe that MIXER converges to local minima that on average lead to better precision and lower recall.
This is likely due to these local minima corresponding to associations that are more conservative, i.e., smaller clusters.

\begin{table}[!t] %
\scriptsize
\centering
\caption{
\mixer\ optimality gap and runtime speedup relative to \mixeropt, determined over 50 Monte Carlo trials.
\mixer\ obtains considerable speedup while maintaining a low optimality gap, indicating a large basin of attraction.
}
\ra{1.2}
\setlength{\tabcolsep}{3pt}
\sisetup{ table-number-alignment=center,
          separate-uncertainty=true,
          table-figures-integer=1,
          table-figures-decimal=1,
          table-figures-uncertainty=1}
\begin{tabular}{ l S c S c@{\hskip 1.6em} S c S[table-figures-decimal=0] c r}
\toprule
$n_o$ & {gap (\%)} && {$p$ (\%)} && {$r$ (\%)} && {runtime (ms)} && speedup \\ \toprule
$2$  &  0.8\pm2.5 && 0.9\pm7.3  && -1.5\pm6.2  && 142\pm 7 && $  0.1\times$ \\
$4$  &  0.5\pm0.7 && 5.6\pm10.8 && -0.6\pm5.6  && 149\pm 7 && $  1.1\times$ \\
$6$  &  0.4\pm0.5 && 5.8\pm12.5 && -2.0\pm8.5  && 161\pm11 && $  6.5\times$ \\
$8$  &  0.4\pm0.3 && 7.8\pm10.5 && -0.7\pm8.9  && 201\pm30 && $ 29.1\times$ \\
$10$ &  0.4\pm0.2 && 5.3\pm11.1 && -2.4\pm10.4 && 218\pm32 && $109.9\times$ \\
$12$ &  0.4\pm0.2 && 9.5\pm26.7 && -3.4\pm9.3  && 252\pm39 && $685.2\times$ \\
\bottomrule
\end{tabular}
\label{tbl:optimality_gap}
\end{table}

\begin{figure}[b]
    \centering
    \includegraphics[clip, trim=4cm 1cm 8cm 1cm, width=1\columnwidth]{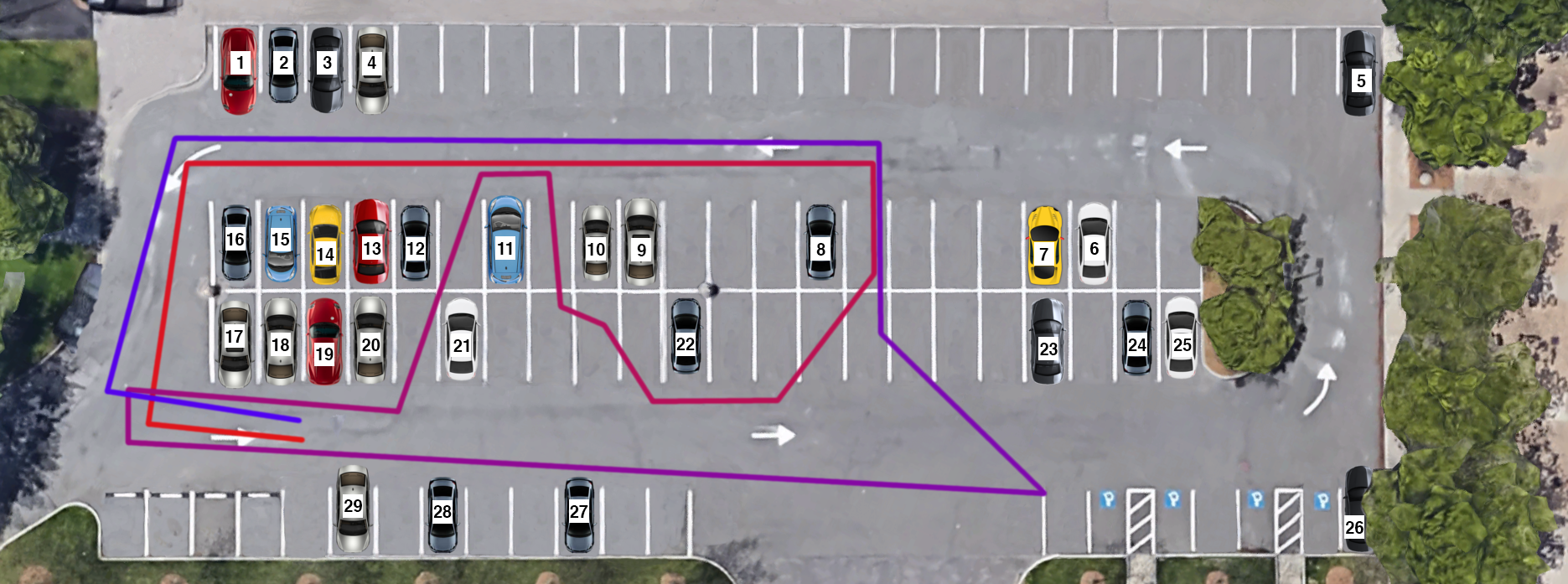}
    \caption{
    Illustration of the parking lot environment.
    Each of the 29 parked cars are located in the parking spots denoted by the numbered car glyph, with approximate color and location.
    The approximate robot path is shown by the red-to-blue gradient path.
    }
	\label{fig:zpark}
\end{figure}

\textbf{Parking Lot Dataset.}
Experimental data is created by driving a Clearpath Jackal fitted with a Velodyne VLP-32 LiDAR and an Intel RealSense D435i
around a parking lot containing 29 cars as shown in Fig.~\ref{fig:zpark}.
The resulting dataset contains RGB frames with time-synchronized robot pose and LiDAR, along with the pre-determined extrinsic calibrations.
Using this dataset, the objective is to associate cars across 100 frames where each view contains noisy, incomplete, and partial detections (i.e., not every car is seen in each frame, and some cars extend out of frame).

\begin{figure}[t!]
    \centering
    \includegraphics[clip, trim=0cm 0.5cm 0cm 3.25cm, width=1\columnwidth]{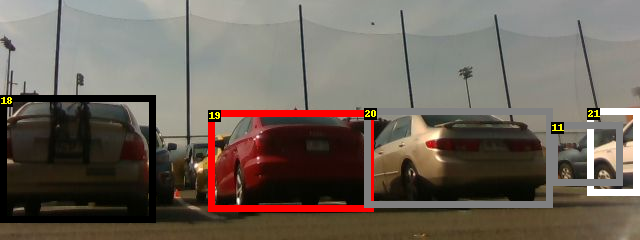}
    \caption{
    Each car's bounding box is colored with its extracted semantic color.
    Harsh lighting in this dataset makes color a weak sensing modality---of the five cars,
    car 18 is misidentified as black, and cars 20 and 11 are deemed inconclusive (grey).
    The bounding box number corresponds to the ground truth label.
    }
	\label{fig:colors}
\end{figure}

\subsubsection{Feature Extraction}
For each RGB frame in the dataset, we use YOLOv3~\cite{redmon2016yolo,redmon2018yolov3} to detect \textit{2D bounding boxes} associated with the 29 cars.
Ground truth associations are generated by manual annotation (see Fig.~\ref{fig:zpark}).
Three additional simple and complementary features of each car are extracted using this bounding box.
A car's \textit{3D centroid} is reconstructed using the median of corresponding LiDAR points that can be reprojected into the 2D bounding box.
The dominant \textit{semantic color} (i.e., red, blue, orange, etc.) is also extracted, with grey representing an inconclusive color.
Finally, the visual appearance of each car is captured using \textit{SIFT keypoints and descriptors}~\cite{lowe1999object,lowe2004sift} from each car's bounding box.

\subsubsection{Computing Association Scores}
Each modality's noisy association matrix $A_{ij}\in[0,1]^{m_i\times m_j}$ (\ie $l=1$) is constructed for each pair of car detections. %
This scoring leverages knowledge about the modality; for example, \textit{bounding box intersection-over-union} can give an indication of similarity for temporally consecutive frames, but for non-consecutive frames this method is inconclusive---we set the similarity score to $0.5$ in this case.
Similarly, similarity based on \textit{spatial proximity} using 3D centroid degrades between temporally distant frames due to odometric drift; therefore, we discount scores such that temporally distant frames yield an inconclusive $0.5$ score.
On the other hand, \textit{color similarity} between two cars is time-independent and takes on $0$ or $1$ depending on semantic color matching;
if either car color is grey, the score is set to $0.5$.
Finally, SIFT matching with Lowe's ratio test~\cite{lowe2004sift} is used to score \textit{visual similarity} of pairs of cars.
Like other modalities, scoring is defined such that weak matching is mapped to the inconclusive $0.5$ score.

Each of the aforementioned similarity scores arise from distinct modalities with their own strengths and weaknesses.
We create a combined modality by formulating diagonal score matrices $S_{ij}$ with $l=4$ (see Section~\ref{sec:formulation}), allowing MIXER to perform multimodal, multiway matching.

\begin{table}[!t] %
\scriptsize
\centering
\caption{
Precision and recall (reported as \texttt{P}/\texttt{R}) of association algorithms for differing modalities.
In each modality column, the algorithm score with the highest $\mathrm{F}_1$ score is bolded.
The highest $\mathrm{F}_1$ score overall is underlined.
}
\ra{1.2}
\setlength{\tabcolsep}{2.5pt}
\begin{tabular}{l c c c c c c c c c c}
\toprule
                                    & bbox        && proximity   && color       && SIFT && combined \\ \toprule
All-pairs                           &  0.47/0.41  &&  \textbf{0.75/0.50} &&  0.11/0.71  && 0.08/1.00 && 0.08/1.00 \\
Consecutive                         &  0.47/0.41  &&  0.93/0.29  &&  \textbf{0.27/0.45} && 0.57/0.43 && 0.68/0.43 \\
Spectral~\cite{Pachauri2013}        &  0.13/0.42  &&  0.15/0.51  &&  0.14/0.45  && 0.29/0.63 && 0.35/0.62 \\
MatchLift~\cite{Chen2014}           &  0.83/0.24  &&  0.83/0.34  &&  0.34/0.17  && 0.51/0.83 && 0.51/0.72 \\
QuickMatch~\cite{tron2017quickmatch}&  0.92/0.20  &&  0.97/0.24  &&  0.18/0.32  && 0.20/0.36 && 0.29/0.34 \\
CLEAR~\cite{fathian2020clear}       &  0.56/0.21  &&  0.14/0.58  &&  0.13/0.41  && 0.30/0.75 && 0.28/0.77 \\
MIXER                               &  \textbf{0.88/0.35} &&  0.85/0.45  &&  0.13/0.35  && \textbf{0.85/0.62}&& \underline{\textbf{0.88/0.82}}\\
\bottomrule
\end{tabular}
\label{tbl:zpark_pr}
\end{table}

\subsubsection{Evaluation}\label{sec:parkeval}
We use precision (\texttt{P}), recall (\texttt{R}), and $\mathrm{F}_1$-score to compare MIXER against state-of-the-art algorithms.
As a baseline, we use an all-pairs association strategy as a naïve multiway matcher, which considers any object pair with an association score greater than $0.5$ a match.
Similarly, we use a consecutive association strategy which performs this simple thresholding, but does not form associations between temporally non-consecutive detections---a commonly used paradigm in e.g., object tracking.
We include the following multiway matching algorithms in our comparison: the Spectral~\cite{Pachauri2013} algorithm, extended by Zhou et al.~\cite{Zhou2015}; MatchLift~\cite{Chen2014} which is based on a convex relaxation with a similar Frobenius objective to MIXER; recent graph clustering algorithms QuickMatch~\cite{tron2017quickmatch} and CLEAR~\cite{fathian2020clear}.

\begin{figure}[t!]
    \centering
    \includegraphics[clip, trim=1.7cm 0.0cm 2.9cm 1.2cm, width=1\columnwidth]{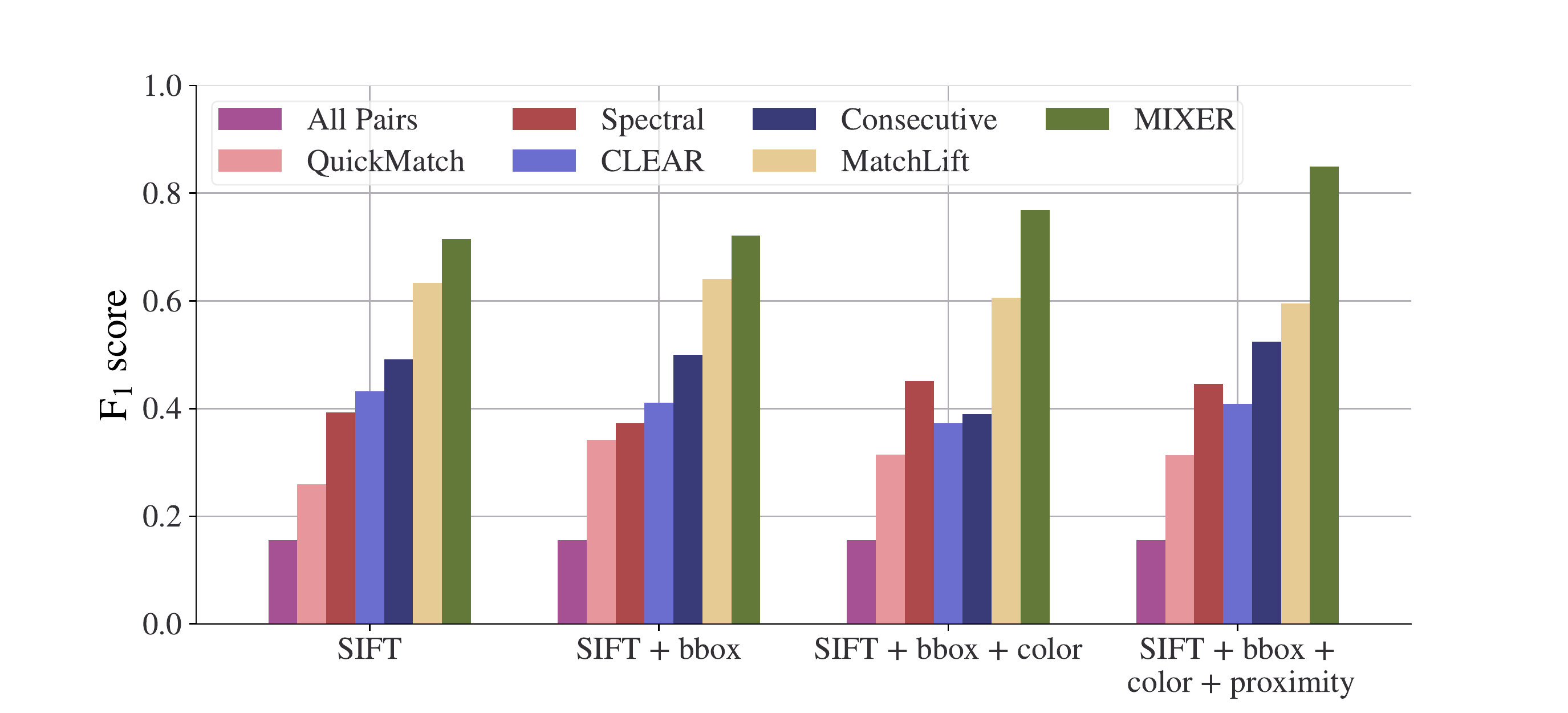}
    \caption{
    Algorithm results starting with visual similarity, the strongest modality, and incrementally combining with others.
    Leveraging the unique information in each modality, MIXER is able to improve upon visual similarity $\mathrm{F}_1$ score by nearly $20\%$.}
	\label{fig:f1ablation}
\end{figure}

Table~\ref{tbl:zpark_pr} lists the \texttt{P}/\texttt{R} results for each association algorithm operating on a given modality.
As expected, the color matching modality is the weakest. %
The bounding box modality also obtains low scores, specifically in \texttt{R}.
This highlights the complementary nature of these two modalities---bounding box similarity is inherently frame-to-frame and thus performs well at identifying matches in consecutive frames (\textuparrow\texttt{P}), but is unable to cluster associations in a multiway matching sense, resulting in fragmented tracks~\cite{milan2016mot16} (\textdownarrow\texttt{R}).
On the other hand, color similarity is time-independent, allowing more non-consecutively seen cars to be matched (\textuparrow\texttt{R}), although semantic colors may not be rich enough to distinguish cars of similar color (\textdownarrow\texttt{P}).
This reasoning also explains why the consecutive algorithm scores the highest $\mathrm{F}_1$ score with the color modality---cars of the same semantic color are more likely to be the same car when the observations are made consecutively.
Visual similarity and proximity produce the best scores, owing to SIFT's robustness even across non-consecutive frames and the local consistency of consecutive position estimates, even in the presence of global drift.
These four modalities are combined and we see that MIXER is capable of recovering associations with high, well-balanced \texttt{P}/\texttt{R}, resulting in an $\mathrm{F}_1$ score of 0.85.
Second-best performance is achieved by MatchLift with an $\mathrm{F}_1$ score of 0.60.

The high \texttt{P}/\texttt{R} of MIXER using multiple modalities is enabled by the $0.5$ ``inconclusive'' score.
By appropriately designing the scoring function of each modality, MIXER is able to delay association decisions that a single modality may have deemed inconclusive.
The benefit is that MIXER \textit{improves} the combined modality associations, whereas other algorithms  struggle to appropriately combine information.
To show that MIXER improves the results of any one modality, we apply the association algorithms to an aggregate association matrix of incrementally combined modalities and show the resulting $\mathrm{F}_1$ scores in Fig.~\ref{fig:f1ablation}.
These results reveal that although the visual similarity modality is strong on its own, MIXER is able to improve it by nearly $20\%$ because of its ability to mix multiple modalities of varying strengths.

\balance %

\bibliographystyle{IEEEtran}
\bibliography{refs}

\begin{thebibliography}{10}
\providecommand{\url}[1]{#1}
\csname url@samestyle\endcsname
\providecommand{\newblock}{\relax}
\providecommand{\bibinfo}[2]{#2}
\providecommand{\BIBentrySTDinterwordspacing}{\spaceskip=0pt\relax}
\providecommand{\BIBentryALTinterwordstretchfactor}{4}
\providecommand{\BIBentryALTinterwordspacing}{\spaceskip=\fontdimen2\font plus
\BIBentryALTinterwordstretchfactor\fontdimen3\font minus
  \fontdimen4\font\relax}
\providecommand{\BIBforeignlanguage}[2]{{%
\expandafter\ifx\csname l@#1\endcsname\relax
\typeout{** WARNING: IEEEtran.bst: No hyphenation pattern has been}%
\typeout{** loaded for the language `#1'. Using the pattern for}%
\typeout{** the default language instead.}%
\else
\language=\csname l@#1\endcsname
\fi
#2}}
\providecommand{\BIBdecl}{\relax}
\BIBdecl

\bibitem{kuhn1955hungarian}
H.~W. Kuhn, ``The hungarian method for the assignment problem,'' \emph{Naval
  Research Logistics Quarterly}, vol.~2, no. 1‐2, pp. 83--97, 1955.

\bibitem{bertsekas1988auction}
D.~P. Bertsekas, ``The auction algorithm: A distributed relaxation method for
  the assignment problem,'' \emph{Annals of operations research}, vol.~14,
  no.~1, pp. 105--123, 1988.

\bibitem{Pachauri2013}
D.~Pachauri, R.~Kondor, and V.~Singh, ``Solving the multi-way matching problem
  by permutation synchronization,'' in \emph{Advances in neural information
  processing systems}, 2013, pp. 1860--1868.

\bibitem{Maset2017}
E.~Maset, F.~Arrigoni, and A.~Fusiello, ``Practical and efficient multi-view
  matching,'' in \emph{IEEE International Conference on Computer Vision}, 2017,
  pp. 4578--4586.

\bibitem{Leonardos2017}
S.~Leonardos, X.~Zhou, and K.~Daniilidis, ``Distributed consistent data
  association via permutation synchronization,'' in \emph{IEEE International
  Conference on Robotics and Automation}, 2017, pp. 2645--2652.

\bibitem{Zhou2015}
X.~Zhou, M.~Zhu, and K.~Daniilidis, ``Multi-image matching via fast alternating
  minimization,'' in \emph{IEEE International Conference on Computer Vision},
  2015, pp. 4032--4040.

\bibitem{Bernard2018}
F.~Bernard, J.~Thunberg, J.~Goncalves, and C.~Theobalt, ``Synchronisation of
  partial multi-matchings via non-negative factorisations,'' \emph{arXiv
  preprint arXiv:1803.06320}, 2018.

\bibitem{tron2017quickmatch}
R.~Tron, X.~Zhou, C.~Esteves, and K.~Daniilidis, ``Fast multi-image matching
  via density-based clustering,'' in \emph{Proceedings of the IEEE
  International Conference on Computer Vision}, Oct 2017.

\bibitem{fathian2020clear}
K.~{Fathian}, K.~{Khosoussi}, Y.~{Tian}, P.~C. {Lusk}, and J.~P. {How},
  ``{CLEAR: A Consistent Lifting, Embedding, and Alignment Rectification
  Algorithm for Multiview Data Association},'' \emph{IEEE Transactions on
  Robotics}, vol.~36, no.~6, pp. 1686--1703, 2020.

\bibitem{Luo2021motlitrev}
W.~Luo, J.~Xing, A.~Milan, X.~Zhang, W.~Liu, and T.-K. Kim, ``Multiple object
  tracking: A literature review,'' \emph{Artificial Intelligence}, vol. 293, p.
  103448, 2021.

\bibitem{karanam2019reid}
S.~{Karanam}, M.~{Gou}, Z.~{Wu}, A.~{Rates-Borras}, O.~{Camps}, and R.~J.
  {Radke}, ``A systematic evaluation and benchmark for person
  re-identification: Features, metrics, and datasets,'' \emph{IEEE Transactions
  on Pattern Analysis and Machine Intelligence}, vol.~41, no.~3, pp. 523--536,
  2019.

\bibitem{Zheng2020heteroreid}
Z.~Wang, Z.~Wang, Y.~Zheng, Y.~Wu, W.~Zeng, and S.~Satoh, ``Beyond
  intra-modality: A survey of heterogeneous person re-identification,'' in
  \emph{Proceedings of the Twenty-Ninth International Joint Conference on
  Artificial Intelligence, {IJCAI-20}}, C.~Bessiere, Ed.\hskip 1em plus 0.5em
  minus 0.4em\relax International Joint Conferences on Artificial Intelligence
  Organization, 7 2020, pp. 4973--4980.

\bibitem{Lowry2016visualplacerecognition}
S.~{Lowry}, N.~{Sünderhauf}, P.~{Newman}, J.~J. {Leonard}, D.~{Cox},
  P.~{Corke}, and M.~J. {Milford}, ``Visual place recognition: A survey,''
  \emph{IEEE Transactions on Robotics}, vol.~32, no.~1, pp. 1--19, 2016.

\bibitem{qin2021semantic}
C.~Qin, Y.~Zhang, Y.~Liu, and G.~Lv, ``Semantic loop closure detection based on
  graph matching in multi-objects scenes,'' \emph{Journal of Visual
  Communication and Image Representation}, p. 103072, 2021.

\bibitem{burkard2009assignment}
R.~Burkard, M.~Dell’Amico, and S.~Martello, ``Assignment problems,''
  \emph{SIAM}, 2009.

\bibitem{munkres1957algorithms}
J.~Munkres, ``Algorithms for the assignment and transportation problems,''
  \emph{Journal of the society for industrial and applied mathematics}, vol.~5,
  no.~1, pp. 32--38, 1957.

\bibitem{koopmans1957assignment}
T.~C. Koopmans and M.~Beckmann, ``Assignment problems and the location of
  economic activities,'' \emph{Econometrica: journal of the Econometric
  Society}, pp. 53--76, 1957.

\bibitem{lawler1963quadratic}
E.~L. Lawler, ``The quadratic assignment problem,'' \emph{Management science},
  vol.~9, no.~4, pp. 586--599, 1963.

\bibitem{loiola2007survey}
E.~M. Loiola, N.~M.~M. de~Abreu, P.~O. Boaventura-Netto, P.~Hahn, and
  T.~Querido, ``A survey for the quadratic assignment problem,'' \emph{European
  journal of operational research}, vol. 176, no.~2, pp. 657--690, 2007.

\bibitem{sahni1976p}
S.~Sahni and T.~Gonzalez, ``P-complete approximation problems,'' \emph{Journal
  of the ACM}, vol.~23, no.~3, pp. 555--565, 1976.

\bibitem{bazaraa1979exact}
M.~S. Bazaraa and A.~N. Elshafei, ``An exact branch-and-bound procedure for the
  quadratic-assignment problem,'' \emph{Naval Research Logistics Quarterly},
  vol.~26, no.~1, pp. 109--121, 1979.

\bibitem{bazaraa1983branch}
M.~Bazaraa and O.~Kirca, ``A branch-and-bound-based heuristic for solving the
  quadratic assignment problem,'' \emph{Naval research logistics quarterly},
  vol.~30, no.~2, pp. 287--304, 1983.

\bibitem{leordeanu2005spectral}
M.~Leordeanu and M.~Hebert, ``A spectral technique for correspondence problems
  using pairwise constraints,'' in \emph{Proceedings of the IEEE conference on
  computer vision and pattern recognition}, 2005, pp. 1482--1489.

\bibitem{swoboda2017study}
P.~Swoboda, C.~Rother, H.~Abu~Alhaija, D.~Kainmuller, and B.~Savchynskyy, ``A
  study of lagrangean decompositions and dual ascent solvers for graph
  matching,'' in \emph{Proceedings of the IEEE conference on computer vision
  and pattern recognition}, 2017, pp. 1607--1616.

\bibitem{zhao1998semidefinite}
Q.~Zhao, S.~E. Karisch, F.~Rendl, and H.~Wolkowicz, ``Semidefinite programming
  relaxations for the quadratic assignment problem,'' \emph{Journal of
  Combinatorial Optimization}, vol.~2, no.~1, pp. 71--109, 1998.

\bibitem{NIPS2013_647bba34}
F.~Fogel, R.~Jenatton, F.~Bach, and A.~D\textquotesingle~Aspremont, ``Convex
  relaxations for permutation problems,'' in \emph{Advances in Neural
  Information Processing Systems}, vol.~26, 2013.

\bibitem{kezurer2015tight}
I.~Kezurer, S.~Z. Kovalsky, R.~Basri, and Y.~Lipman, ``Tight relaxation of
  quadratic matching,'' in \emph{Computer Graphics Forum}, vol.~34,
  no.~5.\hskip 1em plus 0.5em minus 0.4em\relax Wiley Online Library, 2015, pp.
  115--128.

\bibitem{bernard2018ds}
F.~Bernard, C.~Theobalt, and M.~Moeller, ``Ds*: Tighter lifting-free convex
  relaxations for quadratic matching problems,'' in \emph{Proceedings of the
  IEEE conference on computer vision and pattern recognition}, 2018, pp.
  4310--4319.

\bibitem{Zach2010}
C.~Zach, M.~Klopschitz, and M.~Pollefeys, ``Disambiguating visual relations
  using loop constraints,'' in \emph{IEEE Conference on Computer Vision and
  Pattern Recognition}, 2010, pp. 1426--1433.

\bibitem{Nguyen2011}
A.~Nguyen, M.~Ben-Chen, K.~Welnicka, Y.~Ye, and L.~Guibas, ``An optimization
  approach to improving collections of shape maps,'' in \emph{Computer Graphics
  Forum}, vol.~30, no.~5.\hskip 1em plus 0.5em minus 0.4em\relax Wiley Online
  Library, 2011, pp. 1481--1491.

\bibitem{Chen2014}
Y.~Chen, L.~Guibas, and Q.~Huang, ``Near-optimal joint object matching via
  convex relaxation,'' in \emph{International Conference on Machine Learning},
  ser. Proceedings of Machine Learning Research, vol.~32, no.~2, 22--24 Jun
  2014, pp. 100--108.

\bibitem{Hu2018}
N.~Hu, Q.~Huang, B.~Thibert, and L.~J. Guibas, ``Distributable consistent
  multi-object matching,'' in \emph{IEEE Conference on Computer Vision and
  Pattern Recognition}, 2018, pp. 2463--2471.

\bibitem{Yu2016}
J.-G. Yu, G.-S. Xia, A.~Samal, and J.~Tian, ``Globally consistent
  correspondence of multiple feature sets using proximal gauss--seidel
  relaxation,'' \emph{Pattern Recognition}, vol.~51, pp. 255--267, 2016.

\bibitem{Leonardos2018a}
S.~Leonardos and K.~Daniilidis, ``A distributed optimization approach to
  consistent multiway matching,'' in \emph{IEEE Conference on Decision and
  Control}, 2018, pp. 89--96.

\bibitem{leonardos2020low}
S.~Leonardos, X.~Zhou, and K.~Daniilidis, ``A low-rank matrix approximation
  approach to multiway matching with applications in multi-sensory data
  association,'' in \emph{2020 IEEE International Conference on Robotics and
  Automation}, 2020, pp. 8665--8671.

\bibitem{Yan2016a}
J.~Yan, M.~Cho, H.~Zha, X.~Yang, and S.~M. Chu, ``Multi-graph matching via
  affinity optimization with graduated consistency regularization,'' \emph{IEEE
  transactions on pattern analysis and machine intelligence}, vol.~38, no.~6,
  pp. 1228--1242, 2016.

\bibitem{serlin2020distributed}
Z.~Serlin, G.~Yang, B.~Sookraj, C.~Belta, and R.~Tron, ``Distributed and
  consistent multi-image feature matching via quickmatch,'' \emph{The
  International Journal of Robotics Research}, vol.~39, no. 10-11, pp.
  1222--1238, 2020.

\bibitem{yan2013joint}
J.~Yan, Y.~Tian, H.~Zha, X.~Yang, Y.~Zhang, and S.~M. Chu, ``Joint optimization
  for consistent multiple graph matching,'' in \emph{Proceedings of the IEEE
  international conference on computer vision}, 2013, pp. 1649--1656.

\bibitem{shi2016tensor}
X.~Shi, H.~Ling, W.~Hu, J.~Xing, and Y.~Zhang, ``Tensor power iteration for
  multi-graph matching,'' in \emph{Proceedings of the IEEE conference on
  computer vision and pattern recognition}, 2016, pp. 5062--5070.

\bibitem{Swoboda2019}
P.~Swoboda, D.~Kainmuller, A.~Mokarian, C.~Theobalt, and F.~Bernard, ``A convex
  relaxation for multi-graph matching,'' in \emph{IEEE Conference on Computer
  Vision and Pattern Recognition}, 2019.

\bibitem{durrantwhyte1990sensor}
H.~F. Durrant-Whyte, \emph{Sensor Models and Multisensor Integration}.\hskip
  1em plus 0.5em minus 0.4em\relax New York, NY: Springer New York, 1990, pp.
  73--89.

\bibitem{durrantwhyte2016multisensor}
H.~Durrant-Whyte and T.~C. Henderson, \emph{Multisensor Data Fusion}.\hskip 1em
  plus 0.5em minus 0.4em\relax Cham: Springer International Publishing, 2016,
  pp. 867--896.

\bibitem{castanedo2013review}
F.~Castanedo, ``A review of data fusion techniques,'' \emph{The scientific
  world journal}, vol. 2013, 2013.

\bibitem{bar1993estimation}
Y.~Bar-Shalom and X.-R. Li, ``Estimation and tracking- principles, techniques,
  and software,'' \emph{Norwood, MA: Artech House, Inc, 1993.}, 1993.

\bibitem{kittler1998combining}
J.~Kittler, M.~Hatef, R.~P. Duin, and J.~Matas, ``On combining classifiers,''
  \emph{IEEE transactions on pattern analysis and machine intelligence},
  vol.~20, no.~3, pp. 226--239, 1998.

\bibitem{kuncheva2014combining}
L.~I. Kuncheva, \emph{Combining pattern classifiers: methods and
  algorithms}.\hskip 1em plus 0.5em minus 0.4em\relax John Wiley \& Sons, 2014.

\bibitem{shafer1976mathematical}
G.~Shafer, \emph{A mathematical theory of evidence}.\hskip 1em plus 0.5em minus
  0.4em\relax Princeton university press, 1976, vol.~42.

\bibitem{sentz2002combination}
K.~Sentz, S.~Ferson \emph{et~al.}, \emph{Combination of evidence in
  Dempster-Shafer theory}.\hskip 1em plus 0.5em minus 0.4em\relax Sandia
  National Laboratories Albuquerque, 2002, vol. 4015.

\bibitem{murphy1998dempster}
R.~R. Murphy, ``Dempster-shafer theory for sensor fusion in autonomous mobile
  robots,'' \emph{IEEE Transactions on Robotics and Automation}, vol.~14,
  no.~2, pp. 197--206, 1998.

\bibitem{cheeseman1985defense}
P.~Cheeseman, ``In defense of probability,'' in \emph{Proceedings of the 9th
  International Joint Conference on Artificial Intelligence - Volume 2}, ser.
  IJCAI'85.\hskip 1em plus 0.5em minus 0.4em\relax San Francisco, CA, USA:
  Morgan Kaufmann Publishers Inc., 1985, p. 1002–1009.

\bibitem{liu2018learn}
K.~Liu, Y.~Li, N.~Xu, and P.~Natarajan, ``Learn to combine modalities in
  multimodal deep learning,'' \emph{arXiv preprint arXiv:1805.11730}, 2018.

\bibitem{freund1996adaboost}
Y.~Freund and R.~E. Schapire, ``Experiments with a new boosting algorithm,'' in
  \emph{Proceedings of the Thirteenth International Conference on International
  Conference on Machine Learning}, ser. ICML'96.\hskip 1em plus 0.5em minus
  0.4em\relax San Francisco, CA, USA: Morgan Kaufmann Publishers Inc., 1996, p.
  148–156.

\bibitem{granstrom2011learning}
K.~Granstr{\"o}m, T.~B. Sch{\"o}n, J.~I. Nieto, and F.~T. Ramos, ``Learning to
  close loops from range data,'' \emph{The international journal of robotics
  research}, vol.~30, no.~14, pp. 1728--1754, 2011.

\bibitem{bertsekas1997nonlinear}
D.~P. Bertsekas, ``Nonlinear programming,'' \emph{Journal of the Operational
  Research Society}, vol.~48, no.~3, pp. 334--334, 1997.

\bibitem{gurobi}
G.~O. LLC, ``Gurobi optimizer reference manual,'' 2021.

\bibitem{redmon2016yolo}
J.~Redmon, S.~Divvala, R.~Girshick, and A.~Farhadi, ``You only look once:
  Unified, real-time object detection,'' in \emph{Proceedings of the IEEE
  Conference on Computer Vision and Pattern Recognition}, June 2016.

\bibitem{redmon2018yolov3}
J.~Redmon and A.~Farhadi, ``{YOLOv3: An Incremental Improvement},''
  \emph{arXiv}, 2018.

\bibitem{lowe1999object}
D.~G. Lowe, ``Object recognition from local scale-invariant features,'' in
  \emph{International Conference on Computer Vision}, vol.~2, 1999, pp.
  1150--1157.

\bibitem{lowe2004sift}
------, ``Distinctive image features from scale-invariant keypoints,''
  \emph{International journal of computer vision}, vol.~60, no.~2, pp. 91--110,
  2004.

\bibitem{milan2016mot16}
A.~Milan, L.~Leal-Taix{\'e}, I.~Reid, S.~Roth, and K.~Schindler, ``Mot16: A
  benchmark for multi-object tracking,'' \emph{arXiv preprint
  arXiv:1603.00831}, 2016.

\end{thebibliography}

\end{document}